\documentclass[
]{ceurart}

\usepackage{tcolorbox}
\usepackage{graphicx}
\usepackage{multirow}
\usepackage[normalem]{ulem}
\usepackage{subcaption}
\useunder{\uline}{\ul}{}

\begin{document}

\copyrightyear{2020}
\copyrightclause{Copyright for this paper by its authors.
  Use permitted under Creative Commons License Attribution 4.0
  International (CC BY 4.0).}

\newtcolorbox{mybox}[1]{%
    colbacktitle=gray!30,
    coltitle=black,
    fonttitle=\bfseries,
    colback=gray!30,
    colframe=gray!10,
    sharp corners,
    title=#1}

\conference{HASOC (2021) Hate Speech and Offensive Content Identification in English and Indo-Aryan Languages}

\title{Battling Hateful Content in Indic Languages\\ HASOC '21}

\author[1]{Aditya Kadam}[%
email=aditya.kadam@research.iiit.ac.in,
]
\author[1]{Anmol Goel}[%
email=agoel00@gmail.com,
]
\author[1]{Jivitesh Jain}[%
email=jivitesh.jain@students.iiit.ac.in,
]
\author[2]{Jushaan Singh Kalra}[%
email=jushaan18@gmail.com,
]
\author[1]{Mallika Subramanian}[%
email=mallika.subramanian@students.iiit.ac.in,
]
\author[1]{Manvith Reddy}[%
email=manvith.reddy@students.iiit.ac.in,
]
\author[1]{Prashant Kodali}[%
email=prashant.kodali@research.iiit.ac.in,
]
\author[1]{T.H. Arjun}[%
email=arjun.thekoot@research.iiit.ac.in,
]
\author[1]{Manish Shrivastava}[%
email=m.shrivastava@iiit.ac.in,
]
\author[1]{Ponnurangam Kumaraguru}[%
email=pk.guru@iiit.ac.in,
]

\address[1]{International Institute of Information Technology, Hyderabad, India}
\address[2]{Delhi Technological University, Delhi, India}

\begin{abstract}
  The extensive rise in consumption of online social media (OSMs) by a large number of people poses a critical problem of curbing the spread of hateful content on these platforms. With the growing usage of OSMs in multiple languages, the task of detecting and characterizing hate becomes more complex. The subtle variations of code-mixed texts along with switching scripts only add to the complexity. This paper presents a solution for the HASOC 2021 Multilingual Twitter Hate-Speech Detection challenge by team \textit{PreCog IIIT Hyderabad}. We adopt a multilingual transformer based approach and describe our architecture for all 6 subtasks as part of the challenge. Out of the 6 teams that participated in all the subtasks, our submissions rank 3rd overall. 
\end{abstract}


\maketitle

\section{Introduction}
Dissemination of hateful content on nearly all social media is increasingly becoming an alarming concern. In the research community as well, this is a heavily studied research problem~\cite{spread_diffusion_of_hate_on_osm, analysing_hs_targets, hs_racial_bias_mitigation, vulnerable_community_hs, uncompromised_credibility}. Misconduct such as bullying, derogatory comments based on gender, race, religion, threatening remarks etc. are more prevalent today than ever before. The repercussions that such content can have is profound and can result in increased mental stress, emotional outburst and negative psychological impacts~\cite{psych_effects_of_hate_speech}. Hence, curbing the proliferation of this hate speech is imperative. Furthermore, the massive scale at which online social media platforms function makes it an even more pressing issue, which needs to be addressed in a robust manner. Most online social media platforms have imposed strict guidelines \footnote{\url{https://help.twitter.com/en/rules-and-policies/hateful-conduct-policy}} \footnote{\url{https://transparency.fb.com/en-gb/policies/community-standards/hate-speech/}} \footnote{\url{https://support.google.com/youtube/answer/2801939}} to help prevent the spread of hate. In spite of these platform regulations, the dynamics of user-interaction influence the diffusion of (and hence increase in) hate to a large extent~\cite{spread_diffusion_of_hate_on_osm}. 

The problem of hate speech has been addressed by several researchers, but the rise in multilingual content has added to the complexity of identification of hateful content. Majority of these studies deal with high-resource languages such as English, and only recently have low-resource languages -- such as several Indic Languages -- been more deeply explored~\cite{eval_multilingual_indic_bengali_hindi_urdu}. In a country like India, with multitude of regional languages, the phenomenon of Code Mixing/Switching (wherein linguistic units such as phrases/words of two languages occur in a single utterance) is also pervasive.

In this paper we elucidate our approach in solving the six downstream tasks of hate speech identification and characterization in Indian languages as a part of the  \textit{\href{https://hasocfire.github.io/hasoc/2021/index.html}{ `HASOC '21} Hate Speech and Offensive Content Identification in English and Indo-Aryan Languages'} challenge~\cite{hasoc2021mergeoverview}. Motivated by existing architectures, we curate our own pipeline by fusing fine-tuned transformer based models with additional features to solve this challenge and highlight the different methodologies that were adopted for the three languages -- English, Hindi, Marathi, and Code Mixed Hindi - English. We also make our code, methodology and approach public to the research community.~\footnote{\url{https://github.com/Adi2K/Precog-HASOC-2021}}

\section{Literature Review}

Discerning hateful content on social media is an already tricky problem given the challenges associated with it, for instance disrespectful/abusive words could be censored in text, some expressions may not be inherently offensive, however they can be so in the right context~\cite{challenges_hate_speech_social_media}. Owing to the conversational design of social media wherein users can reply to a given comment (either support, refute or irrelevant to the original message), the build-up of threads in response to a hateful message can also intensify hate even if the reply is not hateful on its own. The evolution of such hate intensity has shown diverse patterns and no direct correlation to the parent tweet which makes the task of hate speech detection more difficult~\cite{hate_intensity_threads}.

Significant amount of research has been conducted to evaluate traditional NLP approaches such as character level CNNs, word embedding based approaches and the myriad of variations with LSTMs (sub-word level, hierarchical, BiLSTMs)~\cite{lstm_code_mix_hate_detection_osm}. Likewise, Machine Learning algorithms including SVMs, K-Nearest Neighbours, Multinomial Naive Bayes (MNB) and their respective performances in multilingual text settings have also been explored~\cite{rani-etal-2020-comparative, eval_multilingual_indic_bengali_hindi_urdu, survey_ml_techniques}. Investigating categories of profane words that are commonly used in hate speech is another non-trivial subtask under the hate detection umbrella, primarily because of the different interpretations of words in different cultures/demographics, adaptation of slangs in newer generations etc~\cite{identifying_categorising_profane_words}. 

In recent times however, with the introduction of Transformer based models and their performance in Natural Language Understanding (NLU) tasks, significant work has been done in order to adapt these for multilingual texts as well to leverage transfer between languages. Models such as XLMR, mBERT, MuRIL, RemBERT have gained much popularity and have shown promising results~\cite{bert, xlmr, khanuja2021muril}. Transfer learning based approaches that leverage performance of high resource languages accompanied with CNN classification heads have also shown significant improvements in capturing hateful content on social media platforms~\cite{bert_based_transfer_learning, cross_lingual_transfer_learning}. Sharing and re-utilizing the model weights learnt whilst training on a corpus for a high resource language can aid the process of training for languages that are still under explored~\cite{multilingual_transfer_learning_shared_weights}. 

\section{Dataset}
\subsection{Dataset \& Task Description}

\textbf{Subtask 1} consisted of data for 3 languages, namely -- English, Hindi and Marathi~\cite{hasoc2021overview, gaikwad2021cross}. For English and Hindi, the task was further subdivided into 2 sub-parts: \textbf{\textit{a)}} Identification of hateful v/s non-hateful content and \textbf{\textit{b)}} Characterizing the kind of hate present in a tweet -- either Profane, Hateful, Offensive or None. The distribution of the different data classes for each of the three languages is shown in Table \ref{tab:subtask1_data_distribution}.


\begin{table*}[htp]
\centering
\caption{Distribution of the HASOC 2021 dataset for each task and its associated subtasks. For each language and task, the corresponding number of tweets per class is shown below. Tweets pertaining to Subtask 1 are in three languages -- English, Hindi, Marathi.}
\label{tab:subtask1_data_distribution}
    \begin{tabular}{cccccccc}
        \toprule
        &\multicolumn{2}{c}{\textbf{Task A}} && \multicolumn{4}{c}{\textbf{Task B}} \\ 
        \cline{2-3}\cline{5-8} \textbf{Language}
        & Non-Hateful  & Hateful  && None  & Offensive  & Hate  & Profane \\
        \midrule
        English  & 1342  & 2501 && 1342  & 1196  & 683  & 622 \\ 
        Hindi  & 3161   & 1433   && 3161 & 654  & 566  & 213   \\ 
        Marathi   & 1205  & 669  && - & - & - & -  \\ 
        Subtask 2  & 2899 & 2841 && -  & -  & -  & -\\ 
        \bottomrule
    \end{tabular}
\end{table*}

The focus of \textbf{Subtask 2} was binary classification: Hate \& Offensive or Non Hate-Offensive. The given dataset distribution for this task is shown in Table \ref{tab:subtask1_data_distribution}. This data was accompanied by the following additions~\cite{hasoc2021ICHCLoverview}:
\begin{itemize}
    \item Tweets are English - Hindi Code Mix sentences, and
    \item Classification should not be based on the tweet alone, but should also account for the context as well.
            \newline For example : Consider that in a tweet thread, tweet A is a reply to tweet B. For classifying tweet A, the model can leverage the information from the parent tweet - tweet B.
\end{itemize}

\begin{figure}[t!]
\centering
\begin{subfigure}[t]{.48\textwidth}
  \centering
  \includegraphics[width=.8\linewidth]{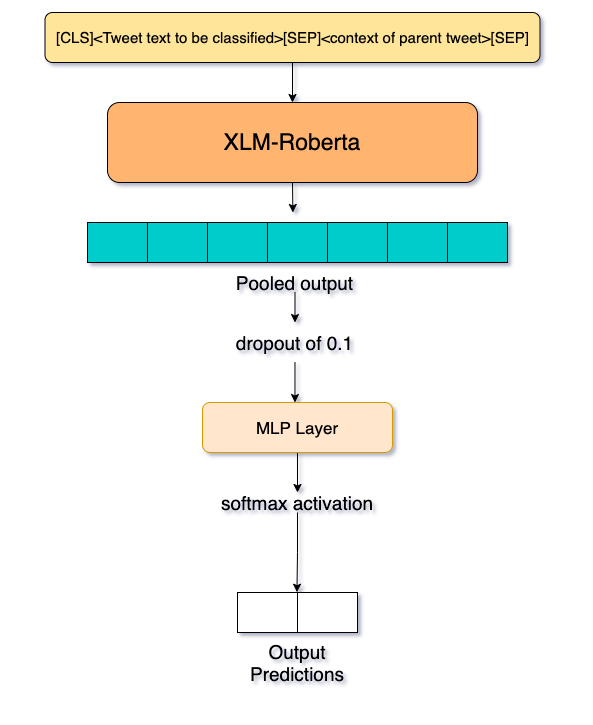}
  \caption{Model pipeline for hate detection in conversational threads for Subtask 2.}
  \label{fig:sub2_archi}
\end{subfigure} \quad
\begin{subfigure}[t]{.48\textwidth}
  \centering
  \includegraphics[width=1\linewidth]{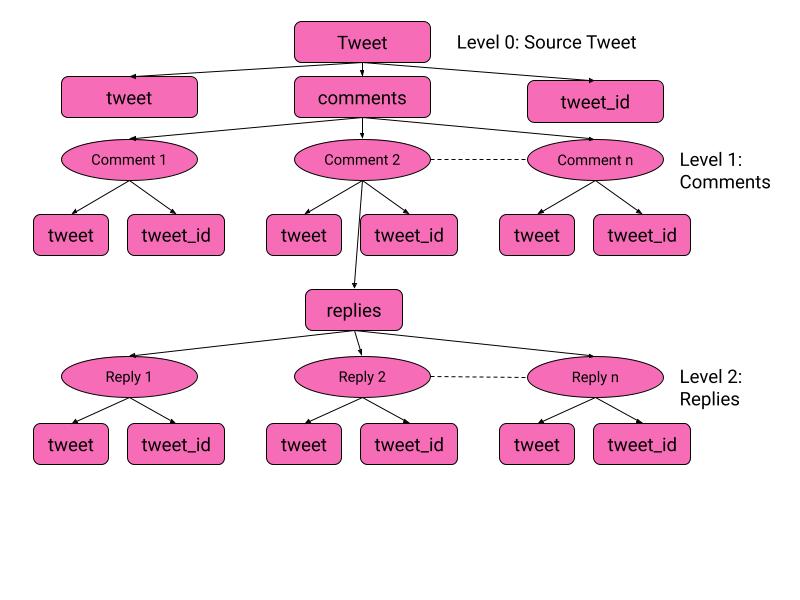}
  \caption{Hierarchy of a conversation thread and its associated comments.}
  \label{fig:sub2_eg}
\end{subfigure}
\caption{Model Pipeline and tweet conversation thread example for Subtask 2.}
\label{fig:sub2_fig}
\end{figure}

Figure \ref{fig:sub2_eg} demonstrates the relationship between the tweets to be classified and their contexts. 

\subsection{Preprocessing Data}

As a precursor to applying any NLP models on text data, we pre-processed the dataset with standard techniques. Given that the data from Twitter is bound to have certain amount of noise and unwanted elements such as -- URLs, mentions etc, these were removed from the tweet texts. Hashtags have a slightly different contribution to analysis of the tweet since they may or may not contribute positively in the classification task. Through the results from our experiments, we observed that omitting the hashtags proved to work better, and hence they were cleaned from the tweet as well. 

Since the data is code mixed, not only in terms of the combination of languages but also with respect to scripts (some English text is written in Roman script, whereas some Hindi text is written in Devanagari apart from Roman), we also normalize the Indic language scripts for Marathi and Hindi. In addition to that, we removed stop words for the Marathi dataset using this list.~\footnote{\url{https://github.com/stopwords-iso/stopwords-mr}} Finally, punctuations were also removed from the dataset texts.

An interesting observation was that for the task of hate detection, the presence of emojis converted to text in the tweets did not improve the performance of our models significantly (rather it reduced the scores by some margin). However, including emojis along with text while classifying hate did have a positive impact since the emoji-text conversion was able to capture hints of sentiment and indirect offensive/profane content.

\section{Methodology}

\subsection{Subtask 1: Identifying Hate, offensive and profane content from the post}

\begin{figure}[t!]
\centering
\begin{subfigure}[t]{.48\textwidth}
  \centering
  \includegraphics[width=.7\linewidth]{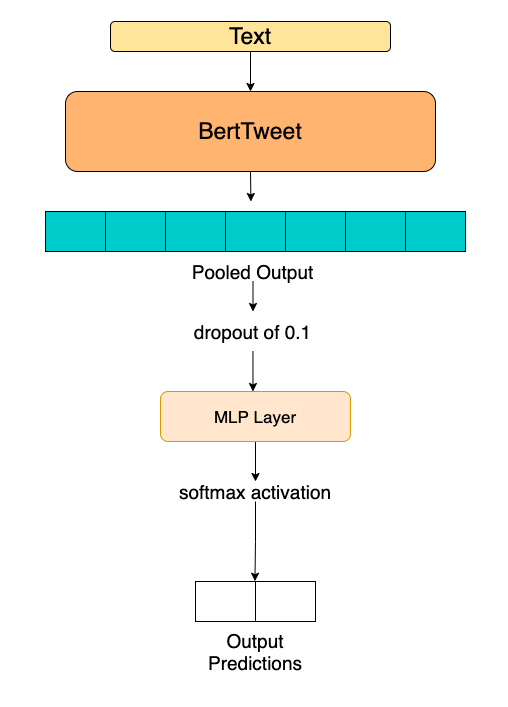}
  \caption{Using BerTweet model with a MLP classifier head.}
  \label{fig:en_twitter}
\end{subfigure}\quad
\begin{subfigure}[t]{.48\textwidth}
  \centering
  \includegraphics[width=.7\linewidth]{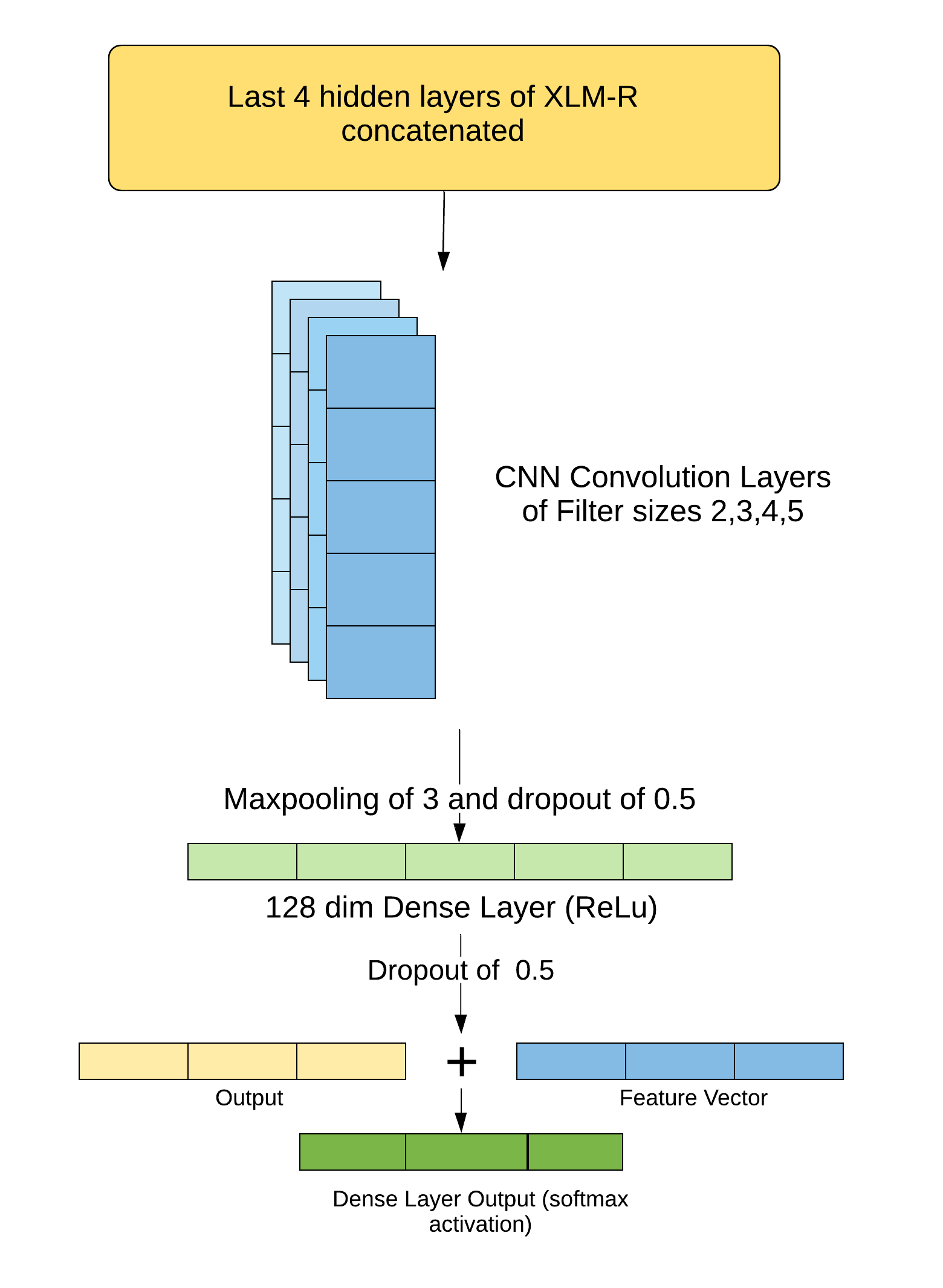}
  \caption{Combining CNN features over XLM-R output and manually generated feature vectors.}
  \label{fig:en_cnn}
\end{subfigure}
\caption{Architecture and pipeline for the models used for the downstream task of hate detection and classification for the English language for Subtask 1.}
\label{fig:eng_archi}
\end{figure}

\subsubsection{English Classifiers}
For the English Subtask 1, the architecture that resulted in the best performance is an ensemble of the following models:
\begin{itemize}
    \item Fine-tuned \textbf{BERTweet} model~\cite{bertweet}
    \item Fine-tuned \textbf{XLM-Roberta}~\cite{xlmr} with CNN Head
\end{itemize}

We use XLM-R, a multilingual model, along with the monolingual model in the ensemble as we found that some of the text in the training set has transliterated Hindi along with some Devanagari text. We extracted textual features such as distribution of ‘?’, ‘!’, capital letters etc. We also use the percentage of profane words and sentiment of the text as a feature. We use profane words list curated from various sources such as words/cuss~\footnote{\url{https://github.com/words/cuss}}, zacanger/profane-words~\footnote{\url{https://github.com/zacanger/profane-words}}, t-davison/lexicons.~\footnote{\url{https://github.com/t-davidson/hate-speech-and-offensive-language/tree/master/lexicons}} For sentiment analysis we use the TweetEval~\cite{barbieri2020tweeteval} model and use its softmax output as a feature to our models.

 Inspired by Kim~\cite{kim2014convolutional} we pass the embedding (concatenated last 4 hidden layers) to a CNN and max-pool convolution layers of various widths to a fully connected layer of size 128 with dropout. We concatenate this 128 dimensional vector with our feature vector. We pass this output onto a dense output layer with softmax activation and cross entropy loss as shown in Figure \ref{fig:eng_archi}.
 
 Along with the previous models, we fine-tune BERTweet, a pre-trained language model for English tweets. BERTweet has the same architecture as BERT and is trained on the pre-training procedure of RoBERTa, but it is trained solely on tweets, thus, making it a viable alternative and suitable for our task. This model has shown state-of-the-art results on tasks based on tweets~\cite{bertweet}. We use the encoder architecture and pass the pooled output through a linear layer for the classification which uses softmax activation and cross-entropy loss as shown in the Figure \ref{fig:eng_archi}.
 
 We also train the models on the previous years datasets but notice that this does not increase the performance of the models but actually degrades the performance in Task 1B due to skewed distribution of classes. Transliteration of emojis didn't improve the performance. The class imbalance in Subtask 1-B degraded the performance of our models hence we tried to improve upon it by using a weighted loss function but we notice that this decreases the performance and that the domain specific distribution is actually helping the models. We also perform K-Fold Validation and use early stopping to avoid over-fitting. We average the probabilities of each class across folds and the two models in our ensemble.
 
\begin{figure}[t!]
\centering
\begin{subfigure}[t]{.48\textwidth}
  \centering
  \includegraphics[width=.88\linewidth]{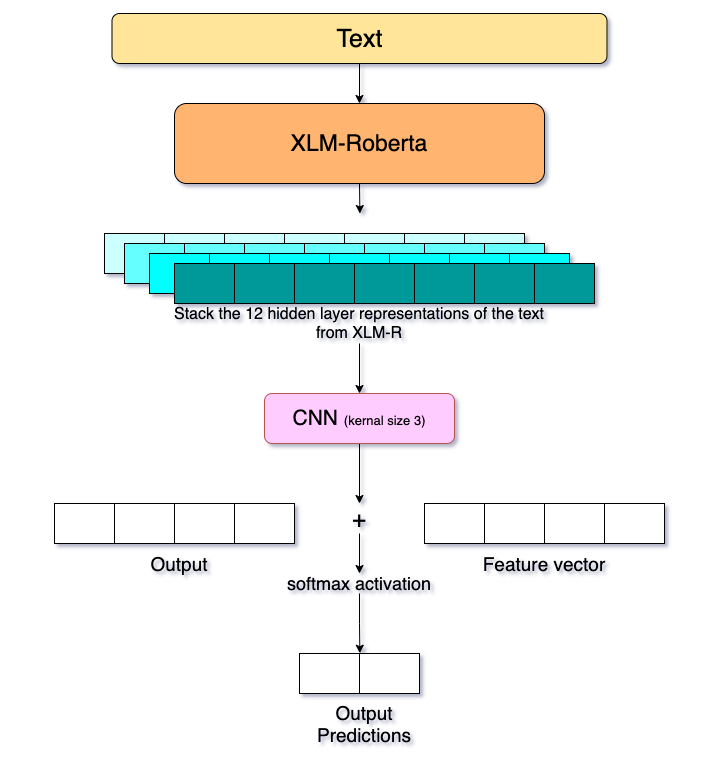}
  \caption{The base architecture for Hindi \& Marathi languages for Subtask 1 using XLM-R with CNN augmented with textual features vector followed by a softmax layer.}
  \label{fig:hi_ma}
\end{subfigure} \quad
\begin{subfigure}[t]{.48\textwidth}
  \centering
  \includegraphics[width=.63\linewidth]{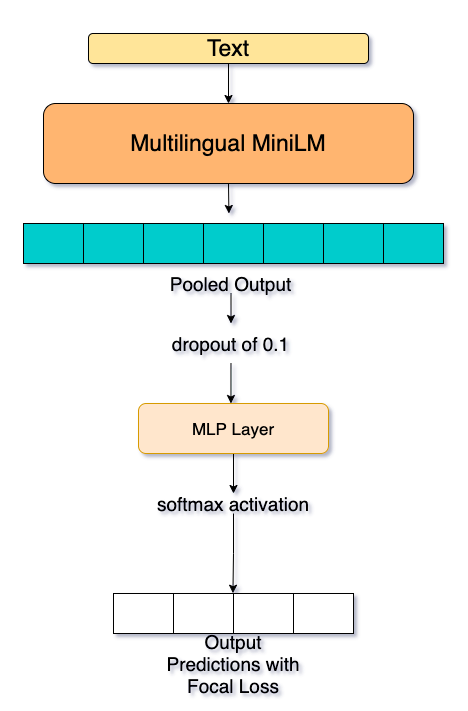}
  \caption{Multilingual MiniLM architecture adopted to overcome class imbalance while characterizing hate for the Hindi Subtask 1-B.}
  \label{fig:hi_mini_lm}
\end{subfigure}
\caption{Architecture and pipeline for the models used for the downstream task of hate detection and classification for the Hindi \& Marathi language for Subtask 1.}
\label{fig:hi_ma_archi}
\end{figure}


\subsubsection{Hindi \& Marathi Classifier}

For both the Hindi and Marathi language, the architecture that performed the best utilized the \textbf{XLM-R} transformer model. This model was able to capture the code-mixed and multilingual nature of the tweets dataset. To amplify the results, we leveraged intermediary representations of the language model as well as textual features that were extracted from the tweets. 
In particular, we utilized the Multilingual MiniLM language model for fine-tuning on Hindi Subtask 1-B. We observed that MiniLM with Focal Loss instead of Cross Entropy Loss performed better than other baselines in the imbalanced multi-class setting of Hindi Subtask 1-B. Focal Loss compensates for class imbalance with a factor that increases the network's sensitivity towards mis-classified samples.

Inspired by Mozafari et al.~\cite{bert_based_transfer_learning} we use the pre-trained representations of the text from 12 hidden layer  of XLM-R model (each of 768 dimensions) and then apply a CNN layer with a kernel size of 3. The output is then passed through a soft-max following which the cross-entropy loss is computed whilst training. This model architecture is represented in Figure \ref{fig:hi_ma_archi}. Tuning hyperparameters such as optimizers, loss functions and dropout layers, we experiment with different options. For the optimizers we try Adadelta and Adam optimizers with Adam working out better. Amongst all loss functions, the Cross Entropy Loss performed the best. As for the dropout layers we explore dropouts in the range 0.1-0.5 and use 0.5 as the final dropout for the model architecture.

We further augment the model features, with two kinds of textual features -- fraction of profane words and sentiment of the tweet. Due to lack of resources for Marathi we catalogue~\footnote{\url{https://github.com/Adi2K/MarathiSwear}} a list of profane words in Marathi and use this to find the fraction of profane words in a tweet. For Hindi, we curate a list of profane words by collating and appending to existing lists~\footnote{\url{https://github.com/neerajvashistha/online-hate-speech-recog/blob/master/data/hi/Hinglish-Offensive-Text-Classification/Hinglish_Profanity_List.csv}}, and use this to score each tweet. As for the sentiment of the tweet, we incorporated off-the-shelf HuggingFace models to obtain the positive, negative and neutral scores for a tweet~\footnote{\url{https://huggingface.co/l3cube-pune/MarathiSentiment}}~\footnote{\url{https://huggingface.co/cardiffnlp/twitter-xlm-roberta-base-sentiment}}. Although the textual features improved the performance for Hindi only by a small margin, for Marathi, manually extracted textual features helped in achieving a significant boost.

For the Marathi Subtask 1, we experimented with a voting ensemble of the \textbf{XLM-Roberta} with CNN Head using the following features:
\begin{itemize}
    \item Word Embedding + Fraction of Profane Words + Sentiment Polarity
    \item Word Embedding + Sentiment Polarity
    \item Word Embedding 
\end{itemize}
However we noticed that the base model with the embedding and the textual features performed better on the leaderboard. 

\subsection{Subtask 2 : Identification of Conversational Hate-Speech in Code-Mixed Languages (ICHCL)}
The tweets for Subtask 2 are code mixed. While the Transformer based encoder models have performed well on various monolingual NLU tasks, their performance does not reach the same level on code mixed sentences. Multilingual transformer based models, have been applied for various code mixed NLU tasks, and have performed better than monolingual transformer based models~\cite{khanuja-etal-2020-gluecos}. For this task, we use XLM-RoBERTa~\cite{xlmr}.
To capture the context and the tweet itself, we modify the input in the following manner, where [CLS] , [SEP] are part of the vocabulary of model, and are used to classify and take multiple sentences as input, respectively.
\\
\begin{tcolorbox}
\textbf{[CLS]} <Tweet text to be classified> \textbf{[SEP]}  <context of parent tweet> \textbf{[SEP]}
\end{tcolorbox}

Here, \textit{<Tweet text to be classified>} is the text of the  tweet/comment/reply that is being classified, while \textit{<Context of parent tweet>} is either just the parent tweet or concatenation of parent tweet and comment, depending on weather the text to be classified is a tweet or a comment or a reply. While classifying a standalone tweet, the context is left empty. The Hindi corpus used to train XLM-Roberta is in Devanagari script, while there is only a small portion of the corpus which is in Romanised form. With the hypothesis that the performance of model will improve if the Hindi tokens are in Devanagari script, we used CSNLI tool~\footnote{https://github.com/irshadbhat/csnli} to convert the Romanised tokens to Devanagari script. However, this normalisation only had a marginal impact on the final performance of the model. We used Huggingface's Trainer API to train the XLM-R model, and the hyperparameters were chosen using the hyperparameter search functionality offered by Trainer API.

\begin{table*}[htp!]
\caption{Performance scores for each of the six subtasks in terms of their test accuracy percentage and Macro F1 scores. All the architectures that were experimented and tested out are tabulated here. We can observe form the results that XLM-R combined with CNN classifier head works best across the languages of Subtask 1, while for Subtask 2, XLM-R with normalised input text performs the best in our experiments.}
\label{tab:perf_scores}
\begin{tabular}{ccccc}
\toprule
    \textit{\textbf{Language}} & \textit{\textbf{Subtask}} & \textit{\textbf{Method}} & \textit{\textbf{Accuracy}} & \textit{\textbf{Macro F1}} \\ 

    \midrule

    \multirow{3}{*}{English} & \multirow{3}{*}{A} & XLM-R + CNN  & 62.30 & 0.62\\
     &  &   Ensemble          & 79.94  &  0.78\\
    & & \textbf{XLM-R + CNN + Sentiment Scores} & \textbf{81.03} & \textbf{0.79} \\ 
    \midrule
    \multirow{2}{*}{English} & \multirow{2}{*}{B} & XLM-R + CNN + Weighted loss & 60.50 & 0.53\\ 
    & &  \textbf{Ensemble} & \textbf{65.18} & \textbf{0.59}\\ 
    
    \midrule
    \multirow{3}{*}{Hindi} & \multirow{3}{*}{A} & MuRIL & 68.90 & 0.60  \\ 
    & & XLM-R Base & 74.00 &  0.76\\ 
    & & \textbf{XLM-R + CNN} & \textbf{80.09} & \textbf{0.77}\\ 
    
    \midrule
    Hindi & B &  \textbf{MiniLM with Focal Loss} & \textbf{72.64} &\textbf{0.51}\\ 
    
    \midrule
    \multirow{3}{*}{Marathi} & \multirow{3}{*}{A} & XLM-R Base & 84.16 & 0.86\\ 
    & & Ensemble & 88.48 & 0.87 \\ 
    & & \textbf{XLM-R + CNN } & \textbf{88.64} & \textbf{0.87} \\ 
    
    \midrule
    \multirow{2}{*}{Hi-En Code mix} & \multirow{2}{*}{2} & XLM-R without norm & 67.58 & 0.67\\ 
    & & \textbf{XLM-R with norm} & \textbf{69.36} & \textbf{0.70} \\ 
    \bottomrule
\end{tabular}
\end{table*}

\subsection{Experiments}

We used Huggingface Transformers~\cite{wolf-etal-2020-transformers} library for implementing the classifiers. For hyper parameter tuning we use Optuna Framework~\footnote{https://optuna.org/} library. Exploring multiple architectures simultaneously, we also tried ensembling an odd number of models following a majority rule based selection. For the English Subtask 1 we also did ensembling with averaged softmax probabilities. However, the increase in complexity of the classification pipeline did not necessarily improve performance scores, considering the size and the distribution of the dataset for Hindi and Marathi but helped in English. Table \ref{tab:perf_scores} captures the Accuracies and F1 scores (corresponding to submissions made on the leaderboard) of all our models for each of the subtasks. 

\section{Conclusion}

In this paper, we presented our approaches for Hate Speech detection on Indian Languages and code mix between Hindi-English using multilingual transformer based encoder models. Although, in this work we have employed different models to address individual language specific subtasks, a multi-task single model based approach, which performs well across all the language pairs, would be an interesting challenge, which we wish to explore as a future work. In addition to this, as part of future work, we would like to improve the performance by carrying out an additional step of domain adaptive pre-training of the  encoder models, and an efficient ensemble of multilingual encoder models. 

\begin{acknowledgments}
  We would like to thank the organisers of \href{https://hasocfire.github.io/hasoc/2021/index.html}{HASOC'21 Shared task} for addressing a crucial problem of hate speech in Indian languages by releasing data resources, and for the smooth conduct of the competition. We would also like to specially thank all members of our research lab, PreCog, for the constructive suggestions during the whole process.
\end{acknowledgments}

\bibliography{sample-ceur}

\end{document}